\documentclass[a4paper,conference]{IEEEtran}
\usepackage{amsmath,graphicx}
\usepackage{amssymb}
\usepackage{color,psfrag}
\usepackage{multirow}
\usepackage{booktabs}
\usepackage{amsfonts} 
\usepackage{listings}
\usepackage{paralist}
\usepackage{amsmath}
\usepackage{verbatim}
\usepackage{enumitem}
\usepackage{footnote}
\usepackage{threeparttable}
\usepackage{etoolbox}
\usepackage{multicol}
\usepackage{tipa}
\usepackage{float}
\usepackage{mathtools}
\usepackage{varioref}
\usepackage{adjustbox}
\usepackage{rotating}
\usepackage{epstopdf}
\usepackage{bm}
\usepackage{caption}
\usepackage{subcaption}
\usepackage{rotating}
\usepackage{mathrsfs}
\usepackage{amsmath}
\usepackage{breqn}

\begin{document}

\title{Quadratic mutual information regularization in real-time deep CNN models}
\author{\IEEEauthorblockN{Maria Tzelepi and Anastasios Tefas \\}
\IEEEauthorblockA{Department of Informatics\\
 Aristotle University of Thessaloniki\\
 Thessaloniki, Greece\\
 Email: $\{$mtzelepi,tefas$\}$@csd.auth.gr}}

\maketitle

\begin{abstract}
In this paper, regularized lightweight deep convolutional neural network models, capable of effectively operating in real-time on devices with restricted computational power for high-resolution video input are proposed. Furthermore, a novel regularization method motivated by the Quadratic Mutual Information, in order to improve the generalization ability of the utilized models is proposed. Extensive experiments on various binary classification problems involved in autonomous systems are performed, indicating the effectiveness of the proposed models as well as of the proposed regularizer.
\end{abstract}


\section{Introduction}
Deep Learning (DL) models, \cite{deng2014tutorial}, and especially deep Convolutional Neural Networks (CNN) have been established among the most efficient research directions in a wide range of computer vision tasks, eclipsing previous shallow algorithms, \cite{gu2018recent}. However, state-of-the-art DL models are usually computation-heavy, obstructing their application on autonomous systems. This has directed the research towards the development of lightweight models capable of running on devices with restricted computational resources such as mobile phones and embedded systems, \cite{howard2017mobilenets,iandola2016squeezenet}. 

\begin{figure}[!h]
\centering
\includegraphics[height=0.134\textheight, width=0.48\textwidth]{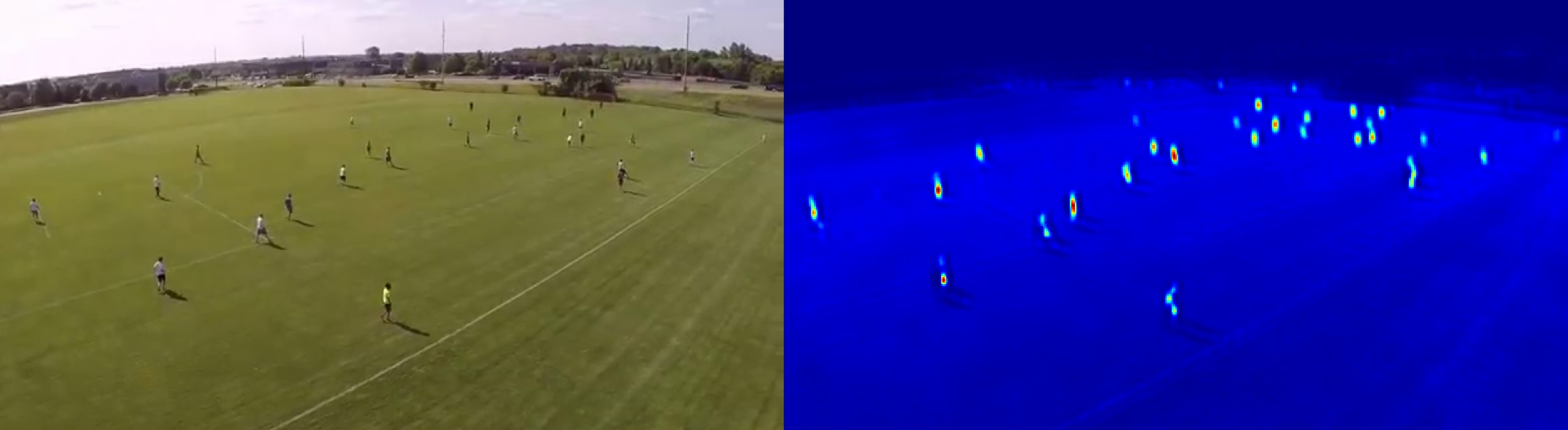}
\caption{An image containing football players and the corresponding predicted heatmap of football player presence.}
\label{heatmap}
\end{figure}

Thus, in this paper, we propose lightweight deep CNN models allowing real-time deployment for high resolution images for specific binary classification problems involved in autonomous robots applications. Specifically, we consider the media coverage of certain events by UAVs (also known as drones). We deal with face, bicycle, and football player detection, as well as crowd detection towards human crowd avoidance. Our goal is to provide semantic heatmaps, by predicting for each location within the captured high-resolution scene the considered object's presence. That is, we train models with RGB input of size either $32 \times 32$  or $64 \times 64$, and then test images are introduced to the network, and, through sliding window process, for every window $32 \times 32$ or $64 \times 64$ respectively, we compute the output of the network at the last convolutional layer.  An example of a football player heatmap is provided in Fig.\ref{heatmap}.

Surveying the relevant literature we can see that several works have emerged towards designing lightweight models. For example, an approach that  proposes to replace $3\times3$ convolutions with $1\times1$ convolutions to create a very small network capable of reducing $50\times$ the number of parameters while obtaining high accuracy is proposed in \cite{iandola2016squeezenet}. However, there is no other work in the recent literature proposing real-time models capable of running on devices with limited computational resources for high resolution input.
Additionally, in \cite{zhang2018shufflenet} where a computation-efficient CNN model is proposed  for mobile devices with limited computing power, it is demonstrated that in order to accomplish real-time deployment, someone has to reduce the input frame resolution to $224 \times224$.

Furthermore, since we deal with lightweight models that usually have inferior performance compared to the more complex ones, we focus on enhancing their performance. That is, the second goal of this work is to propose a novel regularization method in order to circumvent over-fitting and enhance the generalization ability of the proposed real-time models. Generally, this constitutes a major issue in deep learning algorithms, since neural networks are prone to over-fitting due to their high capacity. During the past years, several regularization schemes have been proposed in order to prevent overfitting in neural networks, e.g. common regularization methods, like $\mathscr{L}\emph{1}/\mathscr{L}\emph{2}$ regularization, and Dropout \cite{srivastava2014dropout}. Besides, multitask-learning \cite{caruana1997multitask} has been proposed as a way to improve the generalization ability of a model. For example, in \cite{weston2008deep} the authors introduced techniques developed in semi-supervised learning in the deep learning domain, whilst in \cite{lee2015deeply}, a novel CNN architecture with an SVM classifier at every hidden layer is proposed. This companion objective acts as a kind of feature regularization.

In this work, the so-called \textit{Mutual Information (MI) regularizer} is proposed. The proposed regularizer is inspired by the Quadratic Mutual Information (QMI) measure \cite{torkkola2003feature}, which is a variant of the commonly used Mutual Information, an information-theoretic measure of dependence between random variables. That is, apart from the classification loss, we propose to attach an additional optimization criterion based on the QMI. Recently, QMI reformulated to produce a kernel dimensionality reduction method under the Graph Embedding framework \cite{bouzas}, while in \cite{Passalis_2018_ECCV} a Probabilistic Knowledge Transfer method proposed exploiting the QMI. It is noteworthy that the proposed regularizer is generic and can be applied in several deep learning architectures for classification purposes. 

The remainder of the manuscript is structured as follows. The utilized CNN architectures are described in Section \ref{s1}.
The proposed MI regularizer is presented in Section \ref{s2}. The experiments conducted to validate the proposed method are provided in Section \ref{s3}. Finally, the conclusions are drawn in Section \ref{s4}.

\section{Real-time CNN models}\label{s1}
\begin{figure*}
\begin{subfigure}[t]{0.315\textwidth}
\centering
\includegraphics[width=1.\textwidth, height=0.16\textheight]{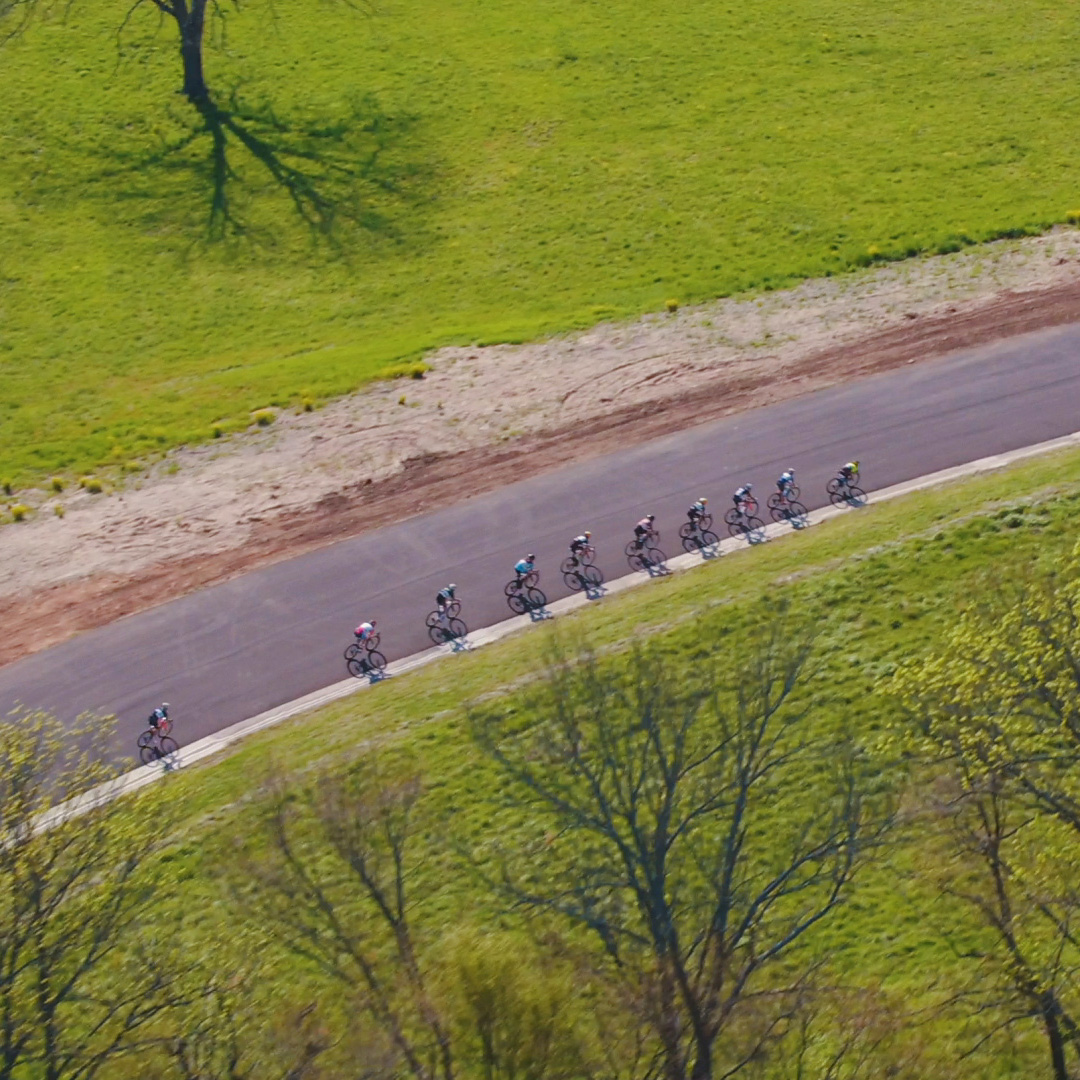}
\caption{}
\label{initial}
  \end{subfigure}
      \begin{subfigure}[t]{0.315\textwidth}
\includegraphics[width=\textwidth, height=0.16\textheight]{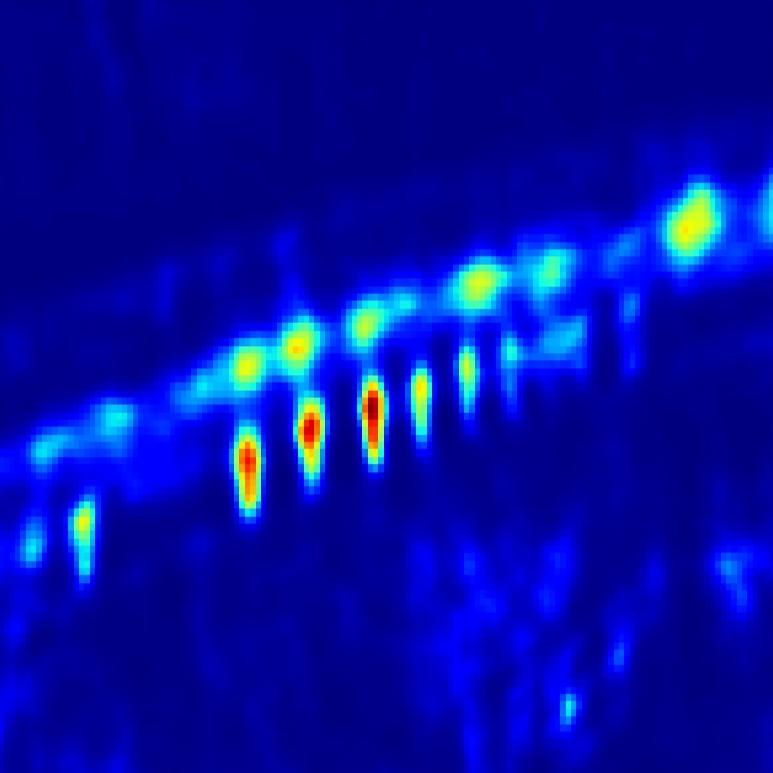}
\caption{}
\label{req3}
  \end{subfigure}
  \begin{subfigure}[t]{0.315\textwidth}
\includegraphics[width=\textwidth, height=0.16\textheight]{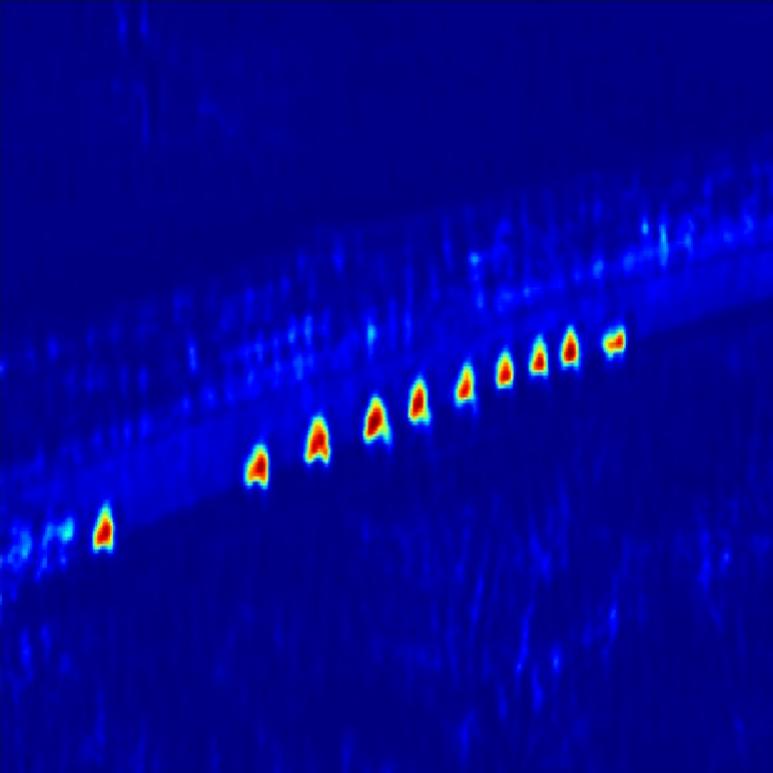}
\caption{}
\label{req5}
  \end{subfigure}
  \caption{An aerial high resolution image containing bicycles (\ref{initial}), and the resulting heatmaps for input of size 640$\times$480 (\ref{req3}) and for input of size 1920$\times$1080 (\ref{req5}) utilizing the proposed VGG-1080p model trained for bicycle detection.}
  \label{req}
  \end{figure*}

In this paper, our goal is to propose effective deep models for various binary classification problems, which allow real-time deployment (about 25 frames per second) on-drone for high resolution images. It should be emphasized that it is of utmost importance for the application to handle high resolution images, since objects in drone-captured images are extremely small, and thus image resizing in order to reach real-time deployment limits, would further shrink the object of interest, rendering the detection infeasible. An example that highlights the demand for high resolution images is provided in Fig. \ref{req}. That is, an aerial image that contains bicycles (bicycles with bicyclists) is provided in Fig. \ref{initial}, and the resulting heatmaps for input of two different resolutions, utilizing the proposed model are provided in Figs. \ref{req3}-\ref{req5}. As it can be observed, as the resolution increases, better performance can be achieved.

\begin{figure}[!h]
\centering
\includegraphics[height=0.11\textheight, width=0.48\textwidth]{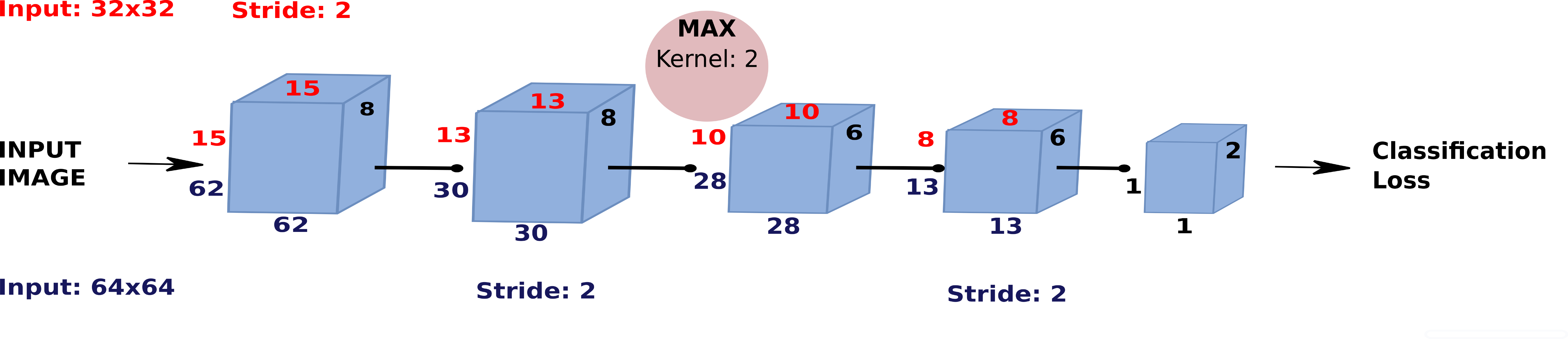}
\caption{VGG-1080p Architecture: Details for input of size 32$\times$32 are printed in red, while details for input of size 64$\times$64 are printed in blue. }
\label{vgg1080pA}
\end{figure}

The objective of this work is two-fold: a) to propose real-time architectures that can be deployed on-drone, and b) to improve the state-of-the-art performance using MI regularization. Thus, we propose a model consisting of only five convolutional layers, by discarding the deepest layers and pruning filters of the widely used VGG-16 model \cite{vgg}. That is, we use the first four convolutional layers of the VGG-16 model with pruned filters, while the last convolutional layer consists of two channels, each for a class, since we deal with binary classification problems. The proposed model runs in real-time on-drone on test images of 1080p (1920$\times$1080) resolution, utilizing the sliding window process, as previously mentioned. The model is abbreviated as as VGG-1080p based on this attribute.  Since we deal with datasets of $32 \times 32$ and $64 \times 64$ input dimensions, we propose two variant models. The models use same kernels and channels, and real-time deployment is achieved with appropriate stride and pooling, as it is shown in Fig. \ref{vgg1080pA}. The evaluation results on the deployment speed for the proposed models are provided in the Experiments Section.

\section{The Proposed MI Regularizer}\label{s2} 
In this paper, a novel regularizer motivated by the Quadratic Mutual Information \cite{torkkola2003feature} is proposed. Apart from the classification loss, we propose a regularization loss derived from the so-called information potentials of the QMI. 
Thus, in this Section, we first introduce the Mutual Information and its quadratic variant, and then we present the proposed MI regularizer. 

We assume a random variable $Y$ representing the image representations of the feature space generated by a specific deep neural layer. We also assume a discrete-value variable $C$ that represents the class labels. For each feature represenation $\bm{y}$ there is a class label $c$. The MI measures dependence between random variables, first introduced by Shannon, \cite{shannon2001mathematical}. That is, the MI measures how much the uncertainty for the class label $c$ is reduced by observing the feature vector $\bm{y}$.  
Let $p(c)$ be the probability of observing the class label $c$, and $p(\bm{y},c)$ the probability density function of the corresponding joint distribution.

The MI between the two random variables is defined as: \\
\begin{equation}
MI(Y,C)= \sum_c \int_{\bm{y}}p(\bm{y},c) \log \frac{p(\bm{y},c)}{p(\bm{y}) P(c)}d\bm{y},
\end{equation}
where $P(c)=\int_{\bm{y}}p(\bm{y},c)d\bm{y}.$ 
MI can also be interpreted as a Kullback-Leibler divergence between the joint probability density $p(\bm{y},c)$ and the product of marginal probabilities $p(\bm{y})$ and $P(c)$.

QMI is derived by replacing the Kullback-Leibler divergence by the quadratic divergence measure \cite{torkkola2003feature}. That is: 
\begin{equation} \label{eq_qmi}
QMI(Y,C)=\sum_c \int_{\bm{y}} \big( p(\bm{y},c) - p(\bm{y})P(c) \big) ^2 d\bm{y}.
\end{equation}
And thus, by expanding eq. (\ref{eq_qmi}) we arrive at the following equation: 

\begin{dmath} \label{eq_qmi2}
QMI(Y,C) = {\sum_c \int_{\bm{y}} p(\bm{y},c) ^2 d\bm{y} +  \sum_c \int_{\bm{y}} p(\bm{y}) ^2 P(c)^2 d\bm{y}} \\
{-2  \sum_c \int_{\bm{y}} p(\bm{y},c)p(\bm{y})P(c) d\bm{y}.}
\end{dmath}

The quantities appearing in eq. (\ref{eq_qmi2}), are called \textit{information potentials} and they are defined as follows: $
V_{IN} = \sum_c \int_{\bm{y}} p(\bm{y},c) ^2 d\bm{y}$, $
V_{ALL} = \sum_c \int_{\bm{y}} p(\bm{y}) ^2 P(c)^2 d\bm{y}$, $
V_{BTW} = \sum_c \int_{\bm{y}} p(\bm{y},c)p(\bm{y})P(c) d\bm{y}$,
and thus, the quadratic mutual information between the data samples and the corresponding class labels can be expressed as follows in terms of the information potentials:
\begin{equation} \label{eq_qmi3}
QMI=V_{IN} + V_{ALL} - 2V_{BTW}.
\end{equation}

If we assume that there are $N_c$ different classes, each of them consisting of $J_p$ samples, the class prior probability for the $c_p$ class is given as: $P(c_p)=\frac{J_p}{N}$, where $N$ corresponds to the total number of samples. Kernel Density Estimation \cite{scott2015multivariate} can be used to estimate the joint density probability: $p(\bm{y},c_p)=\frac{1}{N}\sum_{j=1}^{J_p}K(\bm{y},\bm{y}_{pj};\sigma^2)$, for a symmetric kernel $K$, with width $\sigma$, where we use the notation $\bm{y}_{pj}$ to refer to the $j$-th sample of the $p$-th class, as well as the probability density of $Y$ as $p(\bm{y})=\sum_{p=1}^{J_p}p(\bm{y},c_p)=\frac{1}{N}\sum_{j=1}^NK(\bm{y},\bm{y}_{j};\sigma^2)$.

Thus, eq. (\ref{eq_qmi3}) is formulated as follows: 
\begin{equation}
V_{IN} = \frac{1}{N^2}\sum_{p=1}^{N_c} \sum_{k=1}^{J_p} \sum_{l=1}^{J_p} K(\bm{y}_{pk},\bm{y}_{pl};2\sigma^2),
\end{equation} 

\begin{equation}
V_{ALL} = \frac{1}{N^2}\bigg( \sum_{p=1}^{N_c} \big(\frac{J_p}{N} \big)^2\bigg) \sum_{k=1}^N \sum_{l=1}^N  K(\bm{y}_k,\bm{y}_l;2\sigma^2),
\end{equation} 

\begin{equation}
V_{BTW} = \frac{1}{N^2}\sum_{p=1}^{N_c} \frac{J_p}{N}\sum_{j=1}^{J_p} \sum_{k=1}^N  K(\bm{y}_{pj},\bm{y}_k;2\sigma^2).
\end{equation}

The kernel function $K(\bm{y}_{i},\bm{y}_{j};\sigma^2)$ expresses the similarity between two samples $i$ and $j$. There are several choices for the kernel function, \cite{scott2015multivariate}. For example, in \cite{torkkola2003feature} the Gaussian kernel is used, while in \cite{Passalis_2018_ECCV} the authors utilize a cosine similarity based kernel to avoid defining the width, in order to ensure that a meaningful probability estimation is obtained, since finetuning the width of the kernel is not a straightforward task, \cite{chiu1991bandwidth}. In our experiments, we use as kernel metric a Euclidean based similarity, defined as $K_{ED}= \frac{1}{1 + ||\bm{y}_i-\bm{y}_j||_2^2}$, which also absolves us from defining the width of the kernel. 

The pairwise interactions described above between the samples can be interpreted as follows:
$V_{IN}$ expresses the interactions between pairs of samples inside each class. $V_{ALL}$ expresses the interactions between all pairs of samples, regardless of the  class membership. Finally, $V_{BTW}$ expresses the interactions between samples of each class against all other  samples.

Thus, motivated by the QMI, in this work we propose a novel regularizer in order to enhance the generalization ability of a deep model. That is, apart from the optimization criterion defined by the hinge loss function which aims at separating the samples belonging to different classes, we propose an additional optimization criterion utilizing the information potential defined in eq. (\ref{eq_qmi3}).
We assume that the hinge loss preserves the $V_{BTW}$ information potential which aims to separate samples belonging to different classes. Then, our objective is to maximize pairwise interactions between the samples described by the $V_{IN} + V_{ALL}$ quantities. The derived joint optimization criterion defines an additional loss function, which is attached to the penultimate convolutional layer (that is the last convolutional layer, before the one utilized for the classification task) and acts as regularizer to the main classification objective.

\begin{equation}
J_{MI}=-(V_{IN} + V_{ALL}),
\end{equation} 
where:
\begin{equation}
V_{IN} = \frac{1}{N^2}\sum_{p=1}^{N_c} \sum_{k=1}^{J_p} \sum_{l=1}^{J_p} K_{ED}(\bm{y}_{pk},\bm{y}_{pl}),
\end{equation} 
and 
\begin{equation}
V_{ALL} = \frac{1}{N^2}\bigg( \sum_{p=1}^{N_c} \big(\frac{J_p}{N} \big)^2\bigg) \sum_{k=1}^N \sum_{l=1}^N  K_{ED}(\bm{y}_k,\bm{y}_l).
\end{equation} 

Considering binary classification problems the above optimization criteria can be formulated as follows: \\
\begin{dmath}
V_{IN} =\\
{ \frac{1}{N^2}\sum_{k=1}^{J_1} \sum_{l=1}^{J_1} K_{ED}(\bm{y}_{1k},\bm{y}_{1l})+  \frac{1}{N^2}\sum_{k=1}^{J_2} \sum_{l=1}^{J_2} K_{ED}(\bm{y}_{2k},\bm{y}_{2l}),}
\end{dmath} 
and 
\begin{equation}
V_{ALL} = \frac{1}{N^2}\bigg(\frac{J_1^2+J_2^2}{N^2}\bigg) \sum_{k=1}^N \sum_{l=1}^N  K_{ED}(\bm{y}_k,\bm{y}_l),
\end{equation} 

The total loss for the network training is defined as:
\begin{equation}\label{eqparam}
J_{total}=J_{class}+\eta J_{MI},
\end{equation}
where $J_{class}$ stands for the classification loss, and the parameter $\eta \in[0,1]$ controls the relative importance of $J_{MI}$. We solve the above optimization problem using gradient descent. We should note that the proposed regularizer can be applied for the whole dataset, as well as in terms of mini-batch training. In our experiments we utilize the mini-batch mode. 
We should finally note that in the regularized training we utilize the hinge loss since, as we have experimentally observed, it performs better than the cross entropy one in binary classification problems, however the cross entropy loss could also be utilized.

\section{Experiments}\label{s3}
In this Section, we present the experiments conducted in order to evaluate the proposed models regarding the deployment speed as well as the proposed regularization method. We evaluate the detection speed in terms of frames per second (fps), while we use test accuracy to evaluate the proposed regularizer, since we deal with balanced datasets. Each experiment is executed five times, and the mean value and the standard deviation are reported, considering the maximum value of the test accuracy for each experiment. The probabilistic factor is the random weight initialization.
 
\subsection{Datasets}
In order to evaluate the performance of the proposed regularizer we perform experiments on four datasets constructed for Football Player, Face, Bicycles, and Crowd detection. The Football Player dataset, \cite{tzelepi2019graph}, consists of 98,000 train and 10,000 test images of size $32 \times 32$ that contain equal number of football players and non-football players. The face dataset contains 70,000 train and 7,468 test images of size $32 \times 32$ that contain equal number of face and non-face images. Images of faces have been randomly selected from the AFLW \cite{aflw}, MTFL \cite{mtfl}, and WIDER FACE \cite{wider} datasets. 
The Bicycles dataset, \cite{tzelepi2019graph}, contains 51,200 equally distributed train images of bicycles and non-bicycles, and correspondingly a test set of 10,000 images. Input images are of size $64 \times 64$. Finally, the Crowd-Drone dataset, \cite{tzelepi2019graph}, contains 40,000 UAV-captured train images of equal number of crowded scenes and non-crowded scenes, and 11,550 equally distributed crowded and non-crowded test images. Input images are of size $64 \times 64$. 

\subsection{Implementation Details}
All the experiments conducted using the Caffe Deep Learning framework. 
We use the mini-batch gradient descent for the networks' training. That is, an update is performed for every mini-batch of $N_b$ training samples. The learning rate is set to $10^{-3}$ and drops to $10^{-4}$ gradually, and the batch size is set to 256. The momentum is 0.9. All the models are trained on an NVIDIA GeForce GTX 1080 with 8GB of GPU memory for 100 epochs, and can run in real-time when deployed on an NVIDIA Jetson TX2. In this work, the parameter $\eta$ in (\ref{eqparam}) which controls the relative importance of the proposed regularizer's loss, is set to 0.001, since we have seen that in most cases provides best performance.  Best results are printed in bold.\\
\subsection{Experimental Results}
First, the evaluation results of the proposed models regarding the deployment speed are provided. The performance is tested on a low-power NVIDIA Jetson TX2 module with 8GB of memory, which is a state of the art GPU used for on-board UAV perception. Additionally, in order to accelerate the deployment speed and achieve real-time deployment, TensorRT\footnote{https://developer.nvidia.com/tensorrt} deep learning inference optimizer is utilized. TensorRT is a library that optimizes deep learning models providing FP32 (default) and FP16 optimizations for production deployments of various applications.
In Table \ref{fps} we provide the detection speed in terms of fps for the two proposed models on the NVIDIA Jetson TX2 module without the utilization of the TensorRT optimizer, with the TensorRT on the default mode, and finally with TensorRT on the FP16 mode. As we can see TensorRT and in particular the FP16 mode significantly accelerates the proposed models, achieving detection in-real time for high-resolution images. To gain some intuition about the deployment speed, we note that state-of-the-art detectors run at notably fewer FPS on Jetson TX2, and also for lower resolution input images. For example, YOLO v.2 \cite{yolo} runs at 3.1 fps for input of size $604 \times 604$, while utilizing TensorRT (FP32) runs at 7.8 fps, and further speed up is achieved with the FP16 mode up to 14.4 fps, which remains far away from real-time even for lower input resolution. Finally, we should highlight that the deployment speed regards all the models, that is with and without the proposed regularizer, since the regularizer does not affect the deployment speed.  \\
 	\begin{table}
 	\centering
 	    \resizebox{0.47\textwidth}{!}{ 
	  \begin{tabular}{|c|c|c|c|}
 	   \hline
\textbf{Input} & \textbf{Jetson TX2} &  \textbf{TensorRT-FP32} &  \textbf{TensorRT-FP16}\\ 
\hline
32$\times$32 &  12.3 &  16.9 &  25.7 \\ \hline
64$\times$64 &  8.8 &  18.5 &  25.6 \\ \hline
	  	  \end{tabular}}
	  	  \caption{VGG-1080p: Speed (fps)}
	  	   	   \label{fps}
	    \end{table}

\begin{figure*}
\begin{subfigure}[t]{0.233\textwidth}
\includegraphics[width=\textwidth, height=0.15\textheight]{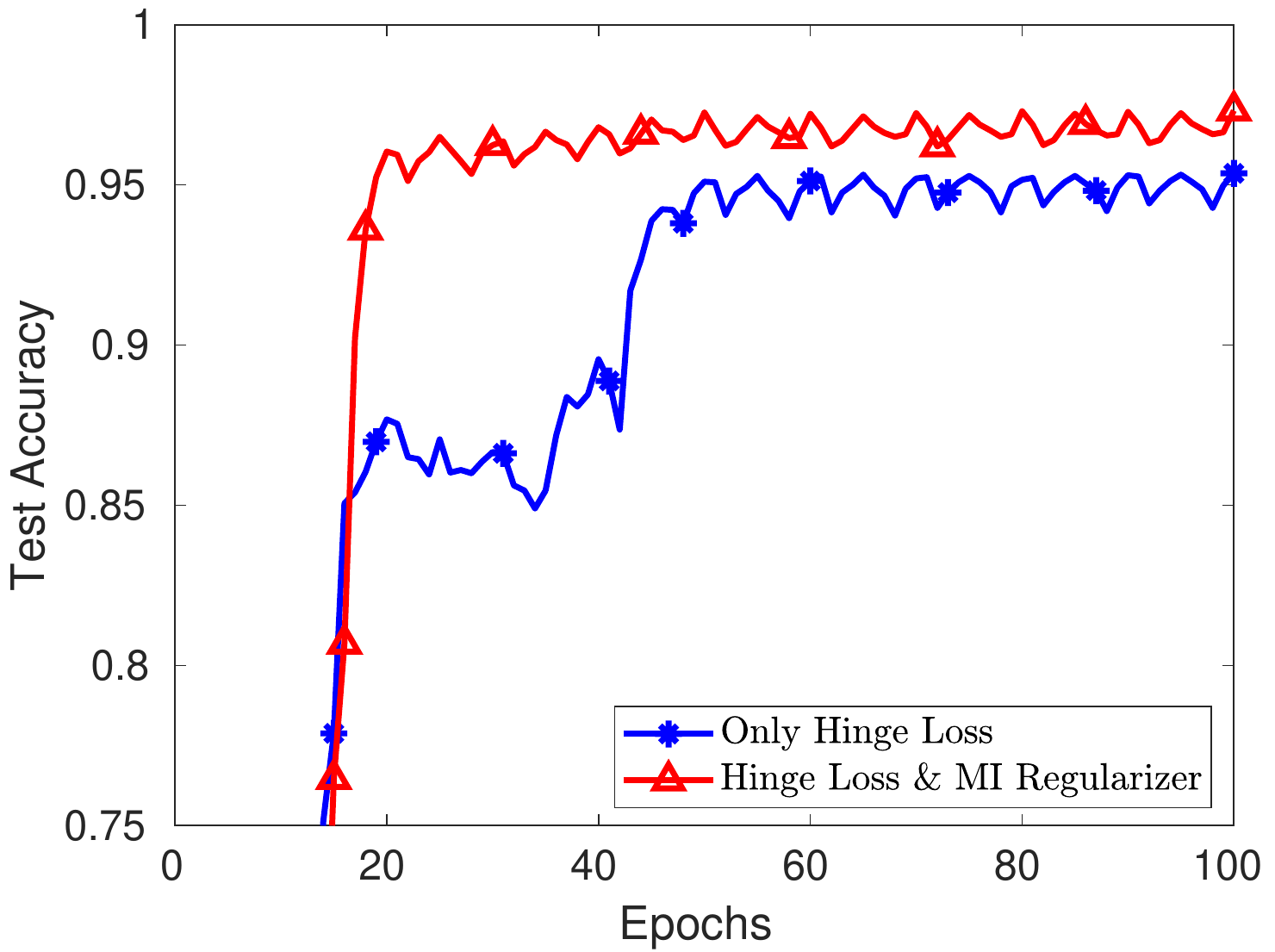}
\caption{Football Player Dataset}
\end{subfigure}
\label{foot-mi-FIG}
\begin{subfigure}[t]{0.233\textwidth}
\includegraphics[width=\textwidth, height=0.15\textheight]{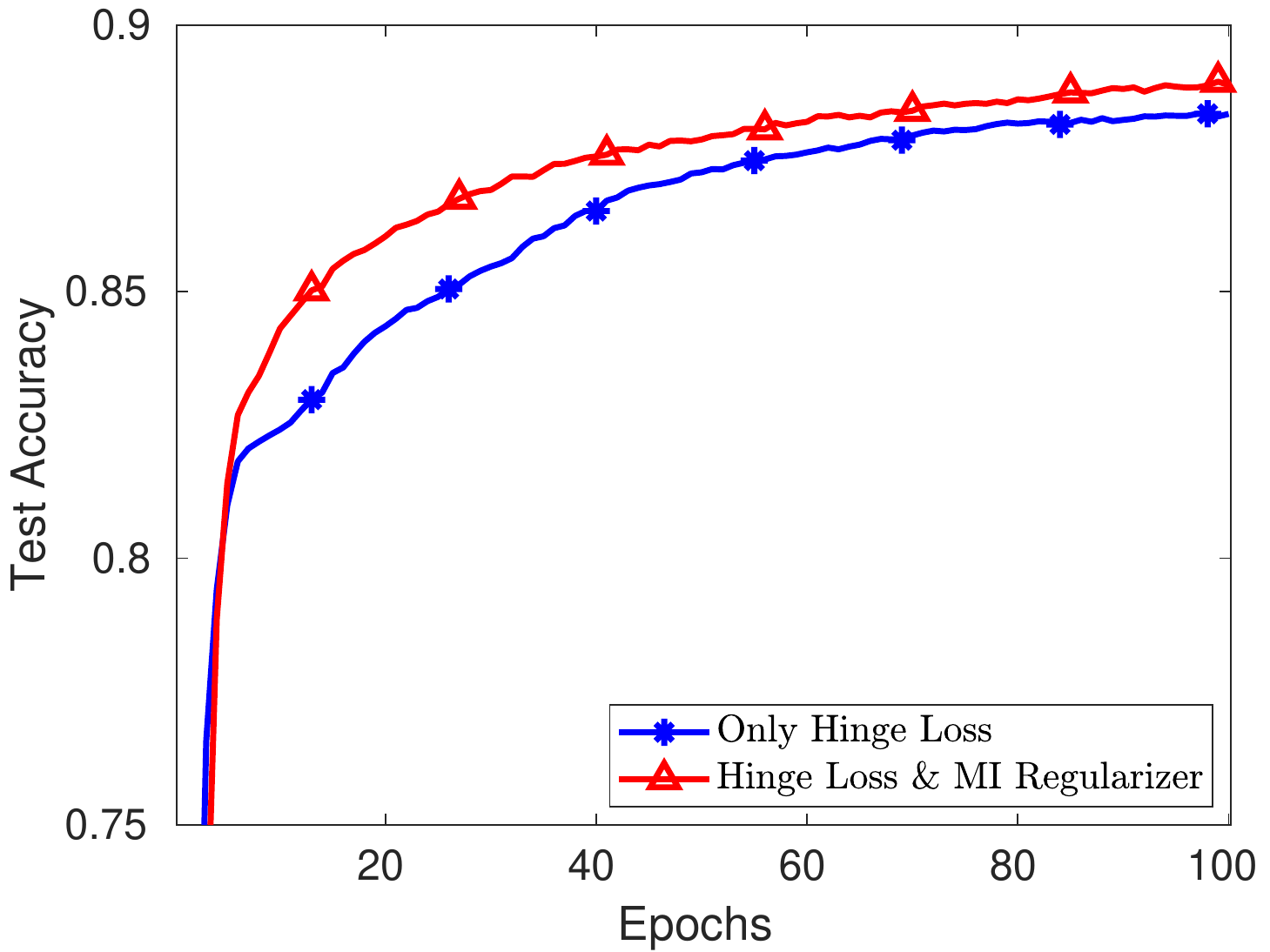}
\caption{Face Dataset}
\end{subfigure}
\label{faces-mi-FIG}
\begin{subfigure}[t]{0.233\textwidth}
\includegraphics[width=\textwidth, height=0.15\textheight]{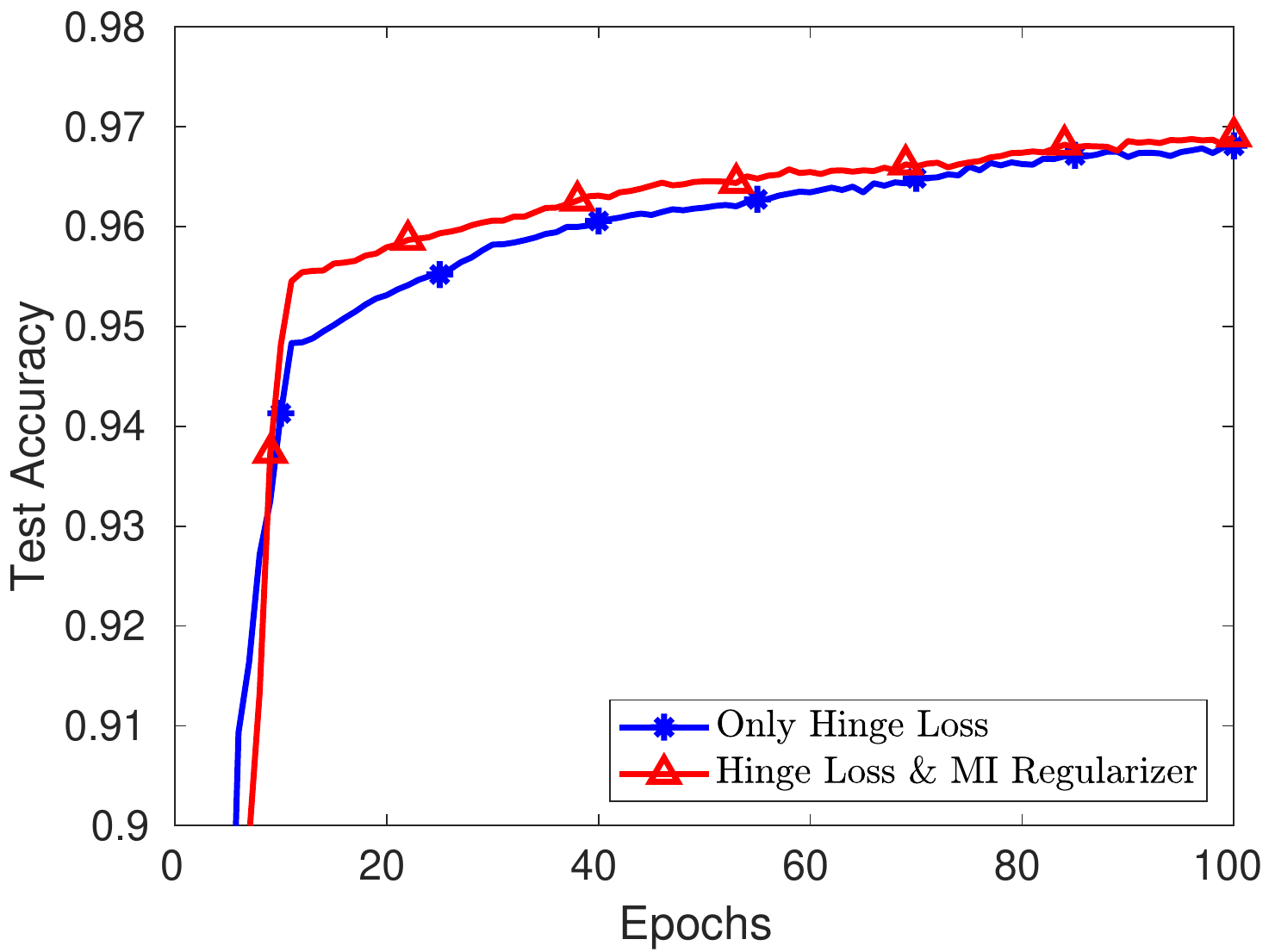}
\caption{Bicycles Dataset}
\end{subfigure}
\label{bikes-mi-FIG}
\begin{subfigure}[t]{0.233\textwidth}
\includegraphics[width=\textwidth, height=0.15\textheight]{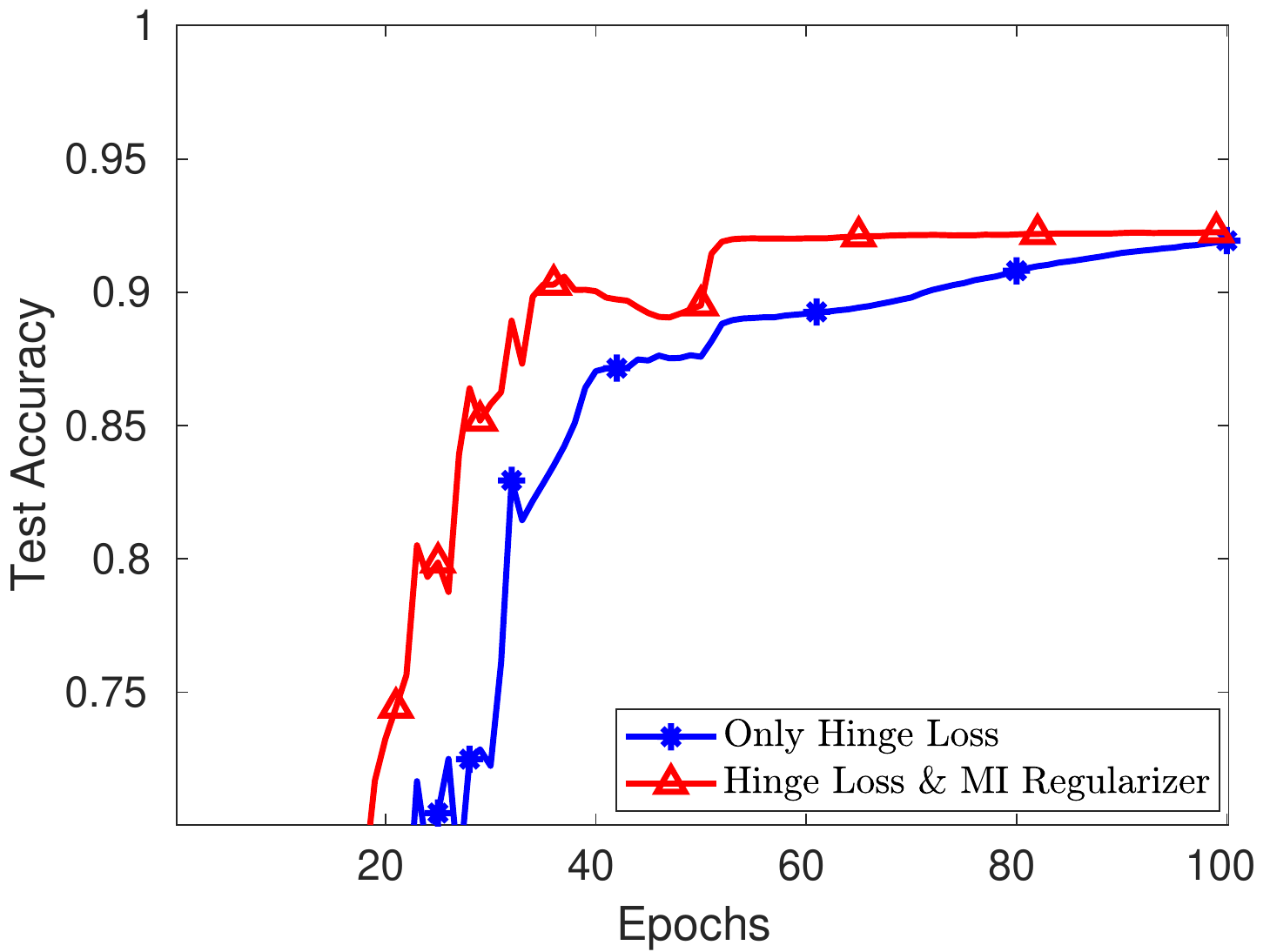}
\caption{Crowd-Drone Dataset}
\end{subfigure}
\caption{MI Regularizer}
\label{mi-FIG}
\end{figure*}

In Table \ref{testacc} we present the mean value and the standard deviation of the test accuracy, for the considered training approaches, that is utilizing only hinge loss, and hinge loss with the proposed MI regularizer. Correspondingly, in Fig. \ref{mi-FIG} we illustrate the curves of mean test accuracy of the only hinge loss training against hinge loss \& MI regularized training. We can see in the demonstrated results, that the proposed MI regularizer remarkably enhances the classification performance on all the utilized datasets.

 	\begin{table}
	    \centering
	  \begin{tabular}{| c | c |  c |}
	       \hline
\textbf{Dataset} &   \textbf{Only Hinge Loss} &\textbf{MI Regularizer} \\ \hline 
Football Player &  0.9568 $\pm$ 0.0100  & \bf{0.9744 $\pm$ 0.0100} \\ \hline 
Face & 0.8841 $\pm$ 0.0040 & \bf{0.8896 $\pm$ 0.0007}\\ \hline 
Bicycles & 0.9684 $\pm$ 0.0037 & \bf{0.9696 $\pm$ 0.0018}\\ \hline 
Crowd - Drone & 0.9194 $\pm$ 0.0082 & \bf{0.9303 $\pm$ 0.0076} \\ \hline 
	  	  \end{tabular}
 	   \caption{Test Accuracy}
 	   \label{testacc}
	    \end{table}

\begin{figure}
\includegraphics[width=0.42\textwidth, height=0.15\textheight]{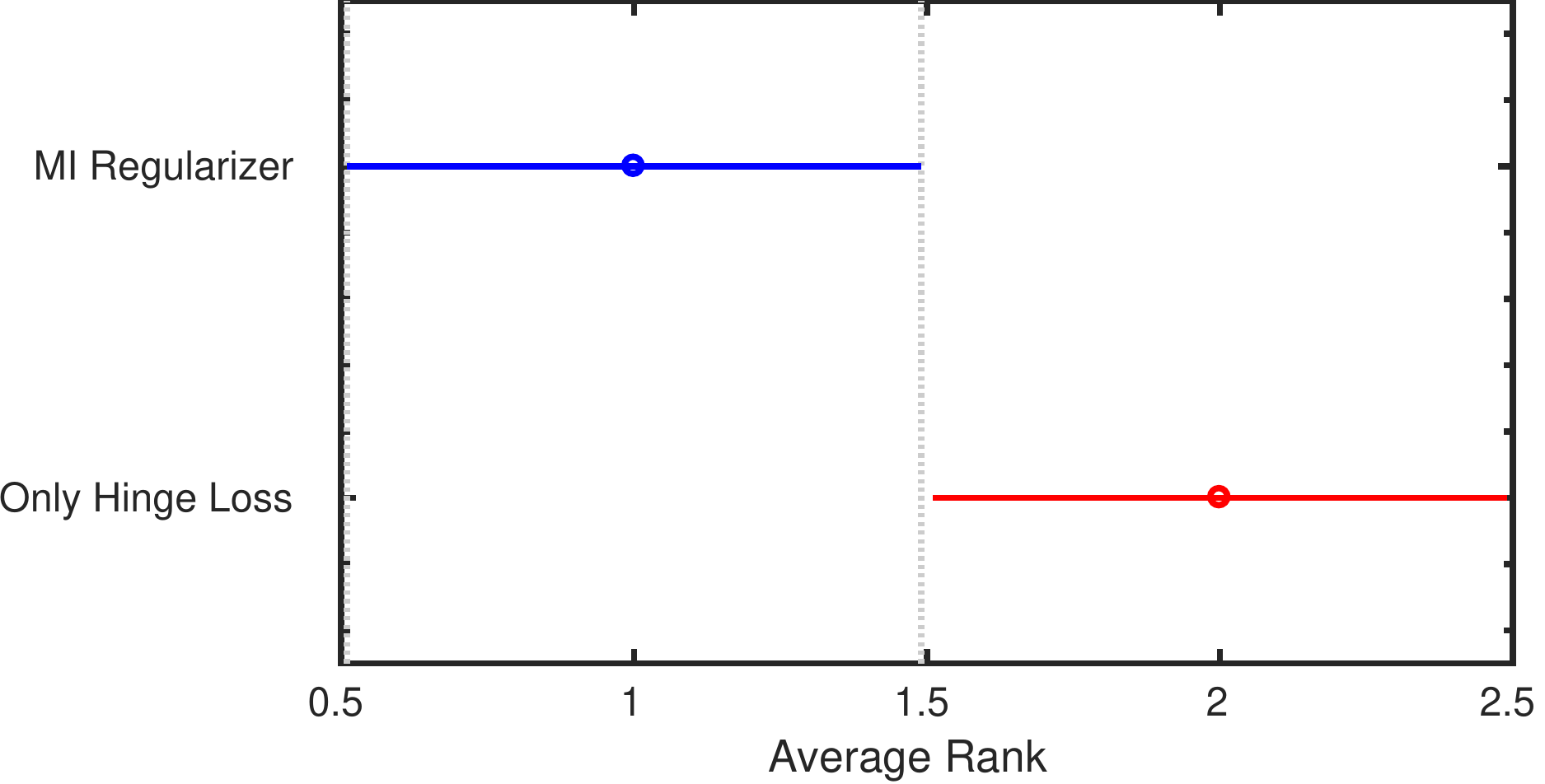}
  \caption{Post-Hoc Bonferroni Test}
  \label{post2}
\end{figure}

Finally, we conducted a post-hoc Bonferroni test \cite{bonfer}, for ranking the proposed regularization method and the only hinge loss training and evaluating the statistical significance of the obtained results. 
The performance of two methods is significantly different, if the corresponding average ranks over the datasets differ by at least the critical difference: 
\begin{equation}
CD = q_a \sqrt{\frac{m(m+1)}{6D}},
 \end{equation}
where m is the number of methods compared, D is the number of datasets and critical values $q_\alpha$ can be found in \cite{bonfer}. In our comparisons
we set $\alpha = 0.05$. 
The number of datasets is four in the performed tests. The compared methods are two, that is the proposed regularizer is compared with a control method which is the only hinge loss training approach. The ranking results are illustrated in Fig. \ref{post2}. 
The vertical axis depicts the two methods, while the horizontal axis depicts the performance ranking. The circles indicate the mean rank and the intervals around them indicate the confidence interval as this is determined by the $CD$ value. Overlapping intervals between two methods indicate that there is not a statistically significant difference between the corresponding ranks, while non-overlapping intervals indicate that the compared methods are significantly different. As we can observe, the proposed regularizer is significantly different against the only hinge loss training approach.

\section{Conclusions}\label{s4}
In this paper, we proposed regularized lightweight CNN models for addressing various binary classification tasks able to run in real-time on-drone. Furthermore, we proposed a novel regularizer motivated by the QMI, the so-called MI regularizer. The performance was evaluated on four datasets. The evaluation results validate the effectiveness of the proposed regularizer in enhancing the generalization ability of the proposed models.

\section*{Acknowledgment}
This project has received funding from the European Union's Horizon 2020 research and innovation programme under grant agreement No 871449 (OpenDR). This publication reflects the authors’ views only. The European Commission is not responsible for any use that may be made of the information it contains.
\bibliographystyle{IEEEtran}
\bibliography{mlsp_tzelepi}

\end{document}